\documentclass[10pt,twocolumn,letterpaper]{article}

\usepackage{cvpr}              

\usepackage[dvipsnames]{xcolor}

\definecolor{cvprblue}{rgb}{0.21,0.49,0.74}
\usepackage[pagebackref,breaklinks,colorlinks,citecolor=cvprblue]{hyperref}

\def\paperID{28} 
\def\confName{CVPR}
\def\confYear{2024}

\title{VDMA: Video Question Answering with Dynamically Generated Multi-Agents}

\author{Noriyuki Kugo \quad\quad Tatsuya Ishibashi \quad\quad Kosuke Ono \quad\quad Yuji Sato\\
Panasonic Connect Co., Ltd.\\
{\tt\small \{kugou.noriyuki,ishibashi.tatsuya001,ono.kosuke,sato.yuji\}@jp.panasonic.com}
}

\begin{document}
\maketitle
\begin{abstract}
This technical report provides a detailed description of our approach to the EgoSchema Challenge 2024.
The EgoSchema Challenge aims to identify the most appropriate responses to questions regarding a given video clip.
In this paper, we propose Video Question Answering with Dynamically Generated Multi-Agents (VDMA).
This method is a complementary approach to existing response generation systems by employing a multi-agent system with dynamically generated expert agents. 
This method aims to provide the most accurate and contextually appropriate responses. This report details the stages of our approach, the tools employed, and the results of our experiments.

\end{abstract}    
\section{Introduction}
\label{sec:intro}

EgoSchema~\cite{egoschema} is an challenging dataset consisting of over 5000 5-choice questions designed for long-form Video Question Answering (VQA) tasks.
The questions in the EgoSchema dataset cover a variety of formats, including inquiries about the purposes of actions, tool usage, and key action detection within the video.
To address this challenge, several studies have been conducted. Approaches that leverage image captioning techniques to generate responses to questions~\cite{Zhang2023ASL} and methodologies utilizing agent-based systems to efficiently extract relevant information~\cite{Wang2024VideoAgentLV} have been proposed. Additionally, recent research in Large Language Models (LLMs) have explored using multi-agent debates to enhance answer accuracy \cite{Du2023ImprovingFA}, and studies involving multi-persona approaches have also been conducted~\cite{wang2023unleashing}. Building on these existing studies, we propose a LLM based multiple expert agents framework for VQA task. Our contributions are summarized as follows:
\begin{itemize}
    \item We propose an LLM based multi-agent VQA system that consists of two stages: dynamic agent generation and question answering by multiple expert agents.
    \item Our proposed method achieved 70.7\% accuracy on the EgoSchema dataset, and experimental results show that our multi-agent approach outperforms single-agent approaches.
\end{itemize}




\section{Our Approach}
\label{sec:formatting}

\hyphenpenalty=10000
\exhyphenpenalty=10000

\begin{figure*}[t]
    \centering
    \includegraphics[width=160mm]{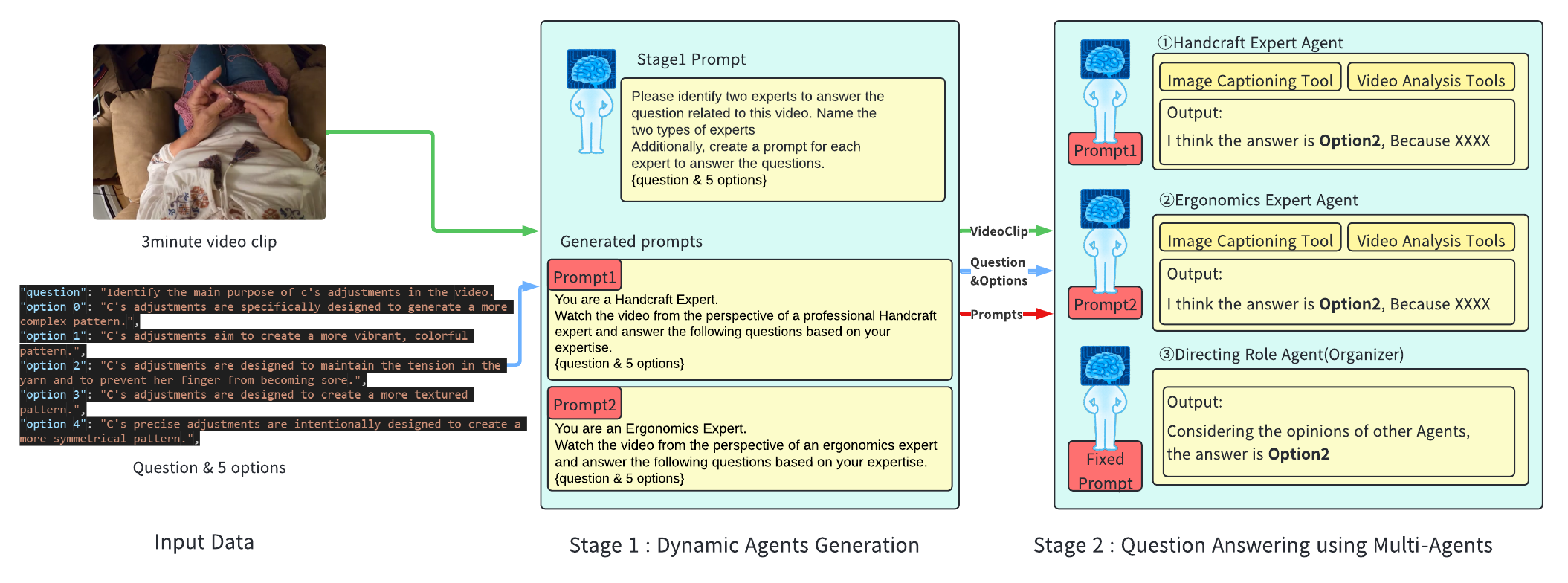}
    \caption{Overall architecture of VDMA}
    \label{fig:overall_architecture}
\end{figure*}

The overall architecture of our system is illustrated in Figure~\ref{fig:overall_architecture}. Our system consists of two stages. In Stage 1, based on the text of the question and the video, we determine the appropriate experts to answer the question and generate prompts for those experts. In Stage 2, we use the generated prompts to create multiple agents. Each generated agent analyzes the video and answers the question using the knowledge of the expert. Finally, an organizer with a predefined prompt consolidates the outputs of each agent and determines the final answer. The following sections provide detailed explanations of Stage 1 and Stage 2.

\subsection{Stage 1 : Dynamic Agent Generation (DAG)}

In Stage 1, agents with appropriate roles are dynamically generated based on the video context and the text of the question. It is widely acknowledged that prompt tuning is a pivotal factor in eliciting more accurate and insightful responses from LLMs. When individuals face challenges that surpass their capacity for resolution, they frequently seek the expertise of specialists. Our methodology emulates this behavior by incorporating DAG within a multi-agent system.
The DAG system analyzes input data, including both the video content and the question text, to dynamically identify the most suitable expert for addressing the question. It then formulates the appropriate prompt to generate that expert agent. Compared to utilizing fixed agent prompts, this method enables the extraction of expert opinions from LLMs that are more contextually aligned with the question text and video context, thereby enhancing the accuracy of the responses.

\subsection{Stage 2 : Question Answering Using Multi-Agent Systems}

In Stage 2, we utilize a multi-agent system to address the questions. The agents in this stage consist of expert agents and an organizer. The expert agents are initialized with the prompts generated in Stage 1. The organizer is responsible for consolidating the opinions of the various expert agents and determining the final answer, using a pre-specified prompt.
Each agent can utilize two tools for analyzing the video and the question text. One tool retrieves image captioning information previously employed in LLoVi~\cite{Zhang2023ASL}. The other tool utilizes GPT-4V for video analysis. Each agent selects the appropriate tool based on the content of the question and the descriptions of the tools. The agents then analyze the video using the chosen tools. Subsequently, they choose the most plausible answer and explain their reasoning. Each agent provides a single response. Finally, the organizer consolidates these responses and determines the final answer.

\section{Experiments}

\subsection{Main Results}

\begin{table*}[t]
  \centering
  \small
  \caption{Models used in the ensemble.}
  \resizebox{170mm}{!}{%
    \begin{tabular}{@{}lcccccc@{}}
      \toprule
      Model & Agent Type & Prompt Generation & Experts' Tools & Organizer's Prompt & Organizer's Tools & Video Analyzer \\
      \midrule
      Model 1 & Single Agent & GPT-4 vision & N/A & default & N/A & N/A \\
      Model 2 & 2 Experts, 1 Organizer & GPT-4 & Captioner, Video Analyzer & default & Captioner, Video Analyzer & GPT-4 vision \\
      Model 3 & 2 Experts, 1 Organizer & GPT-4o & Captioner, Video Analyzer & default & N/A & GPT-4 vision \\
      Model 4 & 3 Experts, 1 Organizer & GPT-4o & Captioner, Video Analyzer & default & N/A & GPT-4o \\
      Model 5 & 3 Experts, 1 Organizer & GPT-4o & Captioner, Video Analyzer & modified & N/A & GPT-4o \\
      \bottomrule
    \end{tabular}%
  }
  \label{tab:main-results-1}
\end{table*}

We evaluated our method on the EgoSchema dataset, which contains a 3-minute video clip, one question about it, and five answer choices. The model selects and outputs the single answer choice that most appropriate answers the question. To improve the accuracy, we applied an ensemble of five models, including our proposed method. Table~\ref{tab:main-results-1} shows the contents of each model.

For the first model, the Azure GPT-4 vision was given a video, questions, and answer choices, and output with the most correct answers. The second model consisted of a multi-agent configuration with two experts and one organizer, and GPT-4 was used to create prompts to each expert. Each expert and organizer was also given Captioner (LaViLa~\cite{Zhao2022LearningVR}) and Video Analyzer (Azure GPT-4 vision) as available tools. The third model was based on the second model, using GPT-4o to create prompts for each expert and not giving the organizer any tools. In addition, the usage instructions of Video Analyzer tool provided to each expert was modified so that it returns the correct or incorrect answer for each of the five choices, rather than returning only the single most accurate choice. The fourth model, based on the third model, increased the number of experts to three and used GPT-4o for Video Analyzer. For the fifth model, the prompt to the organizer was changed so that it would select a shorter, more concise response when unsure of the answer.

\begin{figure*}[th]
    \centering
    \includegraphics[width=135mm]{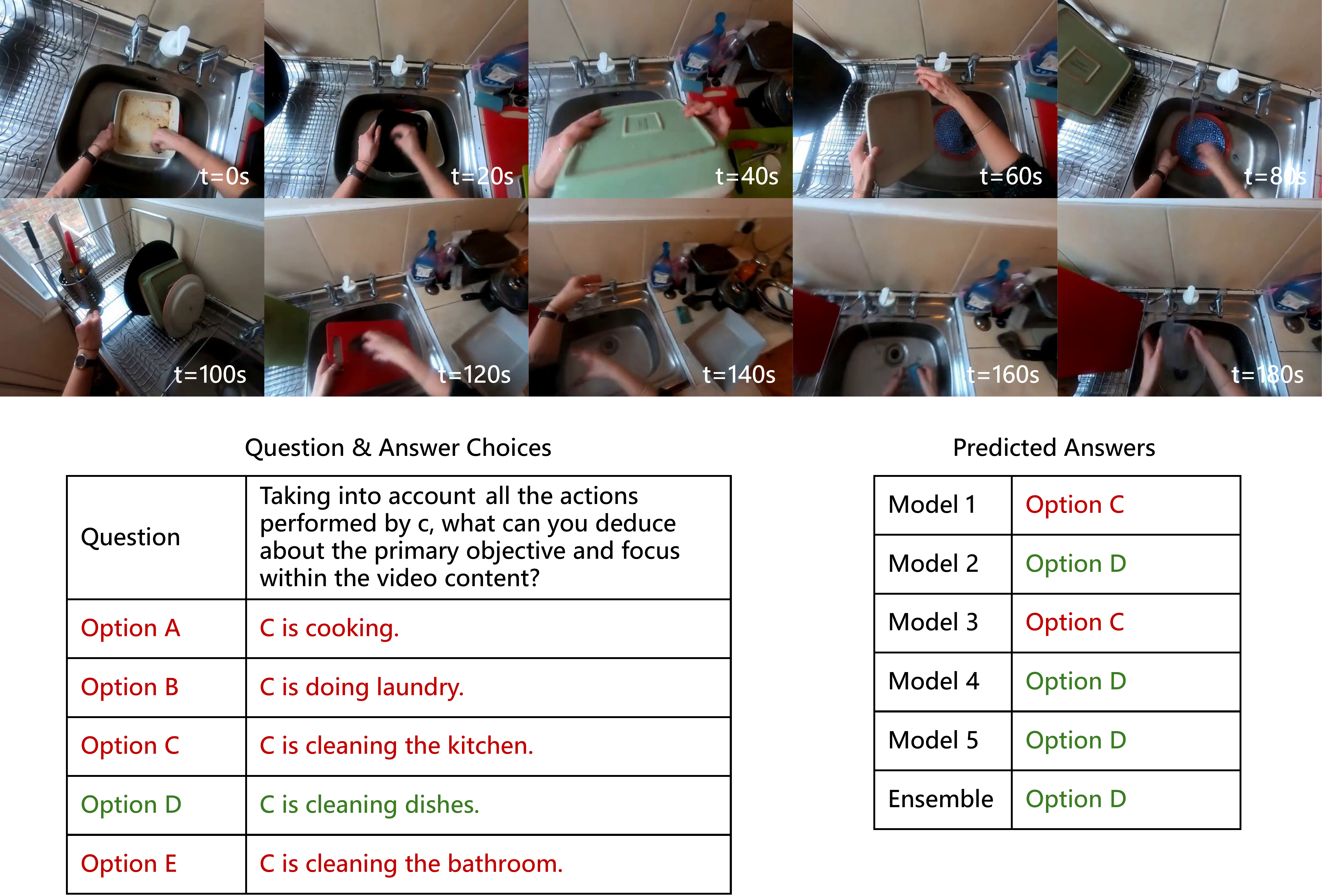}
    \caption{Prediction example on EgoSchema dataset.}
    \label{fig:result}
\end{figure*}

The majority vote was used as the method of ensemble. That is, based on the output of the five models, each answer choice was voted on, and the answer choice with the highest number of votes received was adopted as the final output. In cases where there were multiple first-place votes with the same number of votes for each, the model with the higher individual accuracy was selected. Table~\ref{tab:main-results-2} shows the individual accuracy and the accuracy when the ensemble is applied. Comparing the percentage of correct answers for each individual model, the models with multi-agent configurations (Models 2 to 5) all achieved higher accuracy than the model without such a configuration (Model 1), confirming that the multi-agent configuration we proposed improves accuracy. Also, for the number of agents, higher accuracy was achieved when the number of experts was set to 3 than when the number of experts was set to 2. Comparing Models 4 and 5, instructing the organizer to choose a shorter, more concise response when in unsure of the answer resulted in a 4.9\% decrease in the percentage of correct answers. Next, when the ensemble was applied, a higher percentage of correct answers (70.7\%) was achieved than in any of the five models. Despite the fact that the ensemble method is a relatively simple majority voting method, it was confirmed that it contributes to improved accuracy. Figure~\ref{fig:result} shows an example of the prediction results for each model and ensemble of models on the EgoSchema dataset.

\begin{table}[t]
  \centering
  \caption{Accuracy comparison with individual models and ensemble on EgoSchema Dataset (fullset).}
  \begin{tabular}{@{}lc@{}}
    \toprule
    Model & Acc. (\%) \\
    \midrule
    Model 1 & 62.7 \\
    Model 2 & 63.4 \\
    Model 3 & 62.8 \\
    Model 4 & 68.4 \\
    Model 5 & 63.5 \\
    Ensemble & 70.7 \\
    \bottomrule
  \end{tabular}
  \label{tab:main-results-2}
\end{table}

\subsection{Ablation Study}
This section presents an ablation study to evaluate the effectiveness of different components in our proposed multi-agent VQA method. 
We conducted three additional experiments to examine: 1) performance comparison between multi-agent and single-agent approaches, 2) the impact of domain expert generation in Stage 1, and 3) the effect of the different number of video frames used for analysis.
We used a subset of 500 samples for the additional experiments.

\paragraph{Experiment 1: Multi-Agent vs. Single-Agent} \mbox{} \\
We compared the performance of our multi-agent approach with a single-agent system, which generates the answer in one step. 
The results, as shown in Table~\ref{tab:ablation-1}, indicate a slight improvement in accuracy for the multi-agent approach (73.2\%) compared to the single-agent (72.8\%).
The multi-agent system benefits from the specialized expertise and diverse perspectives of each agent, as they focus on distinct aspects of the video and question. 
Integrating their insights is crucial for resolving ambiguities and refining the final answer, which in turn has the potential to enhance the overall robustness and reliability of the responses.

\paragraph{Experiment 2: Domain Expert vs. AI Assistant} \mbox{} \\
In this experiment, we compared the performance of dynamically generating domain experts in Stage 1 against using uniform AI assistants for all three agents. As shown in Table~\ref{tab:ablation-2}, the approach involving dynamically generated experts achieved higher accuracy (73.2\%) compared to the scenario employing three AI assistants (72.6\%). The domain-specific experts, created dynamically, significantly contributed to solving specialized queries by leveraging their specialized knowledge to provide more accurate and contextually appropriate answers. 
This result suggests the advantage of employing suitable experts over generic AI assistants, particularly in handling complex questions.

\begin{table}[t]
    \centering
    \caption{Performance comparison between multi-agent and single-agent (subset).}
    \begin{tabular}{cc}
        \hline
        Method & Acc. (\%) \\
        \hline
        Single Agent & 72.8 \\
        Multi Agents & 73.2 \\
        \hline
    \end{tabular}
    \label{tab:ablation-1}
\end{table}

\begin{table}[t]
    \centering
    \caption{Performance comparison between domain experts and AI assistants (subset).}
    \begin{tabular}{cc}
        \hline
        Agent Type & Acc. (\%) \\
        \hline
        AI Assistants & 72.6 \\
        Domain Experts & 73.2 \\
        \hline
    \end{tabular}
    \label{tab:ablation-2}
\end{table}

\begin{table}[t]
    \centering
    \caption{Performance comparison between different number of frames per question category (subset). $F$ refers to the number of frames used for video analysis.}
    \resizebox{85mm}{!}{%
        \begin{tabular}{lcccc}
            \hline
            Question Category & Data Ratio & Acc. @$F$=18 & Acc. @$F$=90 \\
            \hline
            Purpose/Goal Identification & 49.2 & 79.3 & 79.3 \\
            Tools and Materials Usage & 21.8 & 72.5 & 76.1 \\
            Key Action/Moment Detection & 21.6 & 63.0 & 64.8 \\
            Action Sequence Analysis & 18.2 & 75.8 & 82.4 \\
            Character Interaction & 9.4 & 74.5 & 70.2 \\
            \hline
            Total & & 73.2 & 75.4 \\
            \hline
        \end{tabular}%
    }
    \label{tab:ablation-3}
\end{table}

\paragraph{Experiment 3: Frame Number Variation} \mbox{} \\
We investigated the impact of varying the number of frames on video analysis performance by comparing 18 frames to 90 frames. As shown in Table~\ref{tab:ablation-3}, using more frames generally improved performance, particularly in \textit{Action Sequence Analysis} with a 6.6\% accuracy increase. This boost is likely due to richer temporal information from additional frames, which helps capture detailed action transitions.

However, \textit{Character Interaction} performance declined with more frames, likely because interactions comprise a smaller video portion, increasing the proportion of non-interaction frames and complicating specific interaction analysis.

The result suggests the need for intelligent frame selection methods that selectively choose frames relevant to the query, potentially reducing noise and enhancing analysis of essential video segments.

\section{Conclusion}


In this technical report, we propose a VDMA for long-form video question answering. Our approach achieved an accuracy of 70.7\% on the EgoSchema dataset, demonstrating that the dynamically generated multi-agent approach is more capable than single-agent solutions by leveraging the expertise and diverse perspectives of multiple agents.

The proposed method involves the use of multiple stages and agents, which increases the computational cost compared to single-agent systems. However, this method leads to more accurate responses, a significant benefit. Furthermore, recent advances in LLM speed and cost reduction are mitigating concerns over computational performance of such systems. Additionally, the DAG approach does not rely on a specific dataset and its versatility allows for easy adaptation to other tasks, suggesting broad applicational potential.

In our study, each agent was limited to a single response per setting, but as indicated in \cite{Du2023ImprovingFA}, adopting a methodology where multi-agent debates continue until a consensus is reached could further enhance accuracy. The selection and optimization of the tools used by agents are directly linked to performance. Future improvements in these tools are expected to further enhance the overall system efficiency and response quality.
{
    \small
    \bibliographystyle{ieeenat_fullname}
    \bibliography{main}

\begin{thebibliography}{6}
\providecommand{\natexlab}[1]{#1}
\providecommand{\url}[1]{\texttt{#1}}
\expandafter\ifx\csname urlstyle\endcsname\relax
  \providecommand{\doi}[1]{doi: #1}\else
  \providecommand{\doi}{doi: \begingroup \urlstyle{rm}\Url}\fi

\bibitem[Du et~al.(2023)Du, Li, Torralba, Tenenbaum, and Mordatch]{Du2023ImprovingFA}
Yilun Du, Shuang Li, Antonio Torralba, Joshua~B. Tenenbaum, and Igor Mordatch.
\newblock Improving factuality and reasoning in language models through multiagent debate.
\newblock \emph{ArXiv}, abs/2305.14325, 2023.

\bibitem[Mangalam et~al.(2024)Mangalam, Akshulakov, and Malik]{egoschema}
Karttikeya Mangalam, Raiymbek Akshulakov, and Jitendra Malik.
\newblock Egoschema: A diagnostic benchmark for very long-form video language understanding.
\newblock \emph{Advances in Neural Information Processing Systems}, 36, 2024.

\bibitem[Wang et~al.(2024)Wang, Zhang, Zohar, and Yeung-Levy]{Wang2024VideoAgentLV}
Xiaohan Wang, Yuhui Zhang, Orr Zohar, and Serena Yeung-Levy.
\newblock Videoagent: Long-form video understanding with large language model as agent.
\newblock \emph{ArXiv}, abs/2403.10517, 2024.

\bibitem[Wang et~al.(2023)Wang, Mao, Wu, Ge, Wei, and Ji]{wang2023unleashing}
Zhenhailong Wang, Shaoguang Mao, Wenshan Wu, Tao Ge, Furu Wei, and Heng Ji.
\newblock Unleashing cognitive synergy in large language models: A task-solving agent through multi-persona self-collaboration.
\newblock \emph{arXiv preprint arXiv:2307.05300}, 2023.

\bibitem[Zhang et~al.(2023)Zhang, Lu, Islam, Wang, Yu, Bansal, and Bertasius]{Zhang2023ASL}
Ce Zhang, Taixi Lu, Md~Mohaiminul Islam, Ziyang Wang, Shoubin Yu, Mohit Bansal, and Gedas Bertasius.
\newblock A simple llm framework for long-range video question-answering.
\newblock \emph{ArXiv}, abs/2312.17235, 2023.

\bibitem[Zhao et~al.(2022)Zhao, Misra, Krahenbuhl, and Girdhar]{Zhao2022LearningVR}
Yue Zhao, Ishan Misra, Philipp Krahenbuhl, and Rohit Girdhar.
\newblock Learning video representations from large language models.
\newblock \emph{2023 IEEE/CVF Conference on Computer Vision and Pattern Recognition (CVPR)}, pages 6586--6597, 2022.

\end{thebibliography}
}


\end{document}